# Improved Part Segmentation Performance by Optimising Realism of Synthetic Images using Cycle Generative Adversarial Networks.

R. Barth[a,*], J. Hemming[a], E.J. Van Henten[b]

[a]*Wageningen University & Research, Greenhouse Horticulture, P.O. Box 644, 6700 AP, Wageningen, The Netherlands*
[b]*Wageningen University & Research, Farm Technology Group, Droevendaalsesteeg 1, 6708 PB, Wageningen, The Netherlands*

**Abstract**

In this paper we report on improved part segmentation performance using convolutional neural networks to reduce the dependency on the large amount of manually annotated empirical images. This was achieved by optimising the visual realism of synthetic agricultural images. In Part I, a cycle consistent generative adversarial network was applied to synthetic and empirical images with the objective to generate more realistic synthetic images by translating them to the empirical domain. We first hypothesise and confirm that plant part image features such as color and texture become more similar to the empirical domain after translation of the synthetic images. Results confirm this with an improved mean color distribution correlation with the empirical data prior of 0.62 and post translation of 0.90. Furthermore, the mean image features of contrast, homogeneity, energy and entropy moved closer to the empirical mean, post translation. In Part II, 7 experiments were performed using convolutional neural networks with different combinations of synthetic, synthetic translated to empirical and empirical images. We hypothesised that the translated images can be used for (i) improved learning of empirical images, and (ii) that learning without any fine-tuning with empirical images is improved by bootstrapping with translated images over bootstrapping with synthetic images. Results confirm our second and third hypotheses. First a maximum intersection-over-union performance was achieved of 0.52 when bootstrapping with translated images and fine-tuning with empirical images; an 8% increase compared to only using synthetic images. Second, training without any empirical fine-tuning resulted in an average IOU of 0.31; a 55% performance increase over previous methods that only used synthetic images. The work presented in this paper can be seen as an important step towards improved sensing for agricultural robotics.

---

*Corresponding author
*URL:* ruud.barth@wur.nl (R. Barth), jochen.hemming@wur.nl (J. Hemming), eldert.vanhenten@wur.nl (E.J. Van Henten)

## 1. Introduction

A key success factor of agricultural robotics performance is a robust underlying perception methodology that can distinguish and localise object parts [1, 13, 2]. In order to train state-of-the-art machine learning methods that can achieve this feat, large annotated empirical image datasets remain required. Synthetic images can help bootstrapping such methods in order to reduce the required amount of annotated empirical data [4]. However, a gap in realism remains between the modelled synthetic images and the empirical ones, plausibly restraining synthetic bootstrapping performance.

The long term objective of our research is to improve plant part segmentation performance. Previous work performed synthetically bootstrapping deep convolutional neural networks (CNN) [4]. In this paper we report on optimising the realism of rendered synthetic images modelled from empirical photographical data [3] that was used in our previous work. We first hypothesise that the dissimilarity between synthetic and empirical images can be qualitatively and quantitatively reduced using unpaired image-to-image translation by cycle-consistent adversarial networks (Cycle-GAN) [29]. Furthermore, we secondly hypothesise that the synthetic images translated to the empirical domain can be used for improved learning of empirical images, potentially further closing the performance gap that remained previously when bootstrapping only with synthetic data [4]. Additionally, our third hypothesis is that without any fine-tuning with empirical images, improved learning of empirical images can be achieved using only translated images as opposed to using only synthetic images.

The key contributions presented in this paper are the (i) further minimisation of the dependency on annotated empirical data for image segmentation learning, and (ii) improving the performance thereof. This can be seen as an important step towards improved sensing for agricultural robotics.

*1.1. Theoretical background*

Convolutional neural networks recently have shown state-of-the-art performance on many image segmentation tasks [9, 24, 6]. However, CNNs require large annotated datasets on a per-pixel level in order to successfully train the large number of free parameters of the deep network. Moreover, in agriculture the high amount of image variety due to a wide range of species, illumination conditions and morphological seasonal growth differences, leads to an increased annotated dataset size dependency. Satisfying this requirement can quickly become a bottleneck for learning.

One solution is to bootstrap CNNs with synthetic images including automatically computed ground truths [10, 27]. Consequently, the bootstrapped network can be fine-tuned with a small set of empirical images, which can result in increased performance over methods without synthetic bootstrapping [4].

Previously we have shown methods to create such a synthetic dataset by realistically rendering 3D modelled plants [3]. Despite intensive manual optimisation for geometry, color and textures, we have shown that a discrepancy remains between the synthetic and empirical images. Although this dataset can be used for successful synthetic bootstrapping and improved empirical learning, there remained a difference between the achieved performance and the theoretical optimal performance [4].

Recently, the advent of generative adversarial networks (GAN) introduced another method of image data generation [15]. In GANs two deep convolutional neural networks are trained simultaneously and adversarially: a generative model $G$ and a discriminative model $D$. The generative model's goal is to capture the feature distribution of a dataset by learning to generate images thereof from latent variables (e.g. random noise vectors). The discriminative model in turn evaluates to what extent the generated image is a true member of the dataset. In other words, model $G$ is optimised to trick model $D$ while model $D$ is optimising to not get fooled by model $G$. In Figure 1 a schematic overview of this learning process is shown. As both models can be implemented as CNNs, the error can be back-propagated to minimise the loss of both models simultaneously. The result after training is a model $G$ that can generate new random images highly similar to the learned dataset. This method is useful if one wants to generate more similar images from the same domain. Given that this does not provide a corresponding ground truth, this method was not pursued for this paper, although as we'll see later, it does provide a fundamental building block towards the method that was used.

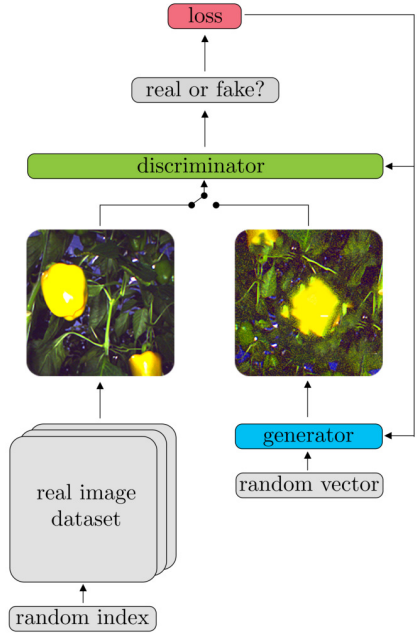

Figure 1: Learning schematic of a generative adversarial network. In each learning step, a discriminator receives either a random real domain image from a database or a translated image from the generator. The discriminator determines if this image is real or translated. Through a loss function, feedback to both the discriminator and the generator is given to optimise their tasks. In this example, the generator learns to synthesise empirical photographic images from random vectors.

In later approaches, GANs were conditioned with an additional input image from another domain [21], forming an image pair that had some relation with each other (e.g. a color image and its label or class mapping). The generator was tasked with image-to-image translation to create a coherent image (e.g. color) from a corresponding pair image (e.g. label map). The discriminator's goal is then to evaluate if input pairs are either real or generated. The loss can then be fed back to both the discriminator and generator to improve on their tasks. The result after training is a generator $G$ that can translate images from one domain X (e.g. color images) to images in another domain Y (e.g. label maps) or more formally notated as $G : X \to Y$. In Figure 2 a schematic overview of the learning process is shown. Given that this does not provide additional novel training pairs, we also did not pursue this method for this paper, although again this methodology provides a useful building block for our work.

A requirement for image-to-image translation using conditional GANs, is that images from both domains are geometrically paired. For our objective of translating images from the synthetic domain to the empirical domain, this requirement was not met because images from both domains did not geometrically correspond one-to-one.

A recent approach aimed to dissolve this paired geometry requirement by investigating unpaired image-to-image translation [29]. In cycle-consistent adversarial networks (Cycle-GAN), a mapping $G : X \to Y$ is learned whilst at the same time also the inverse mapping $F : Y \to X$ is learned. Both domains $X$ and $Y$ have corresponding discriminators $D_X$ and $D_Y$. Hence, $D_X$ ensures $G$ to translate $X$ similar to $Y$ whilst $D_Y$ tries to safeguard a preferably indistinguishable conversion of $Y$ to $X$.

However since the domains are unpaired, the translation at this point does not guarantee that an individual image $x \in X$ is mapped to a geometrically similar image in domain $Y$ (or vice versa $y \in Y$ to $X$). This is because there are boundless mappings from $x$ that result in the same target distribution of $Y$. Therefore the mapping needs to be constrained in a way such that the original geometry is maintained.

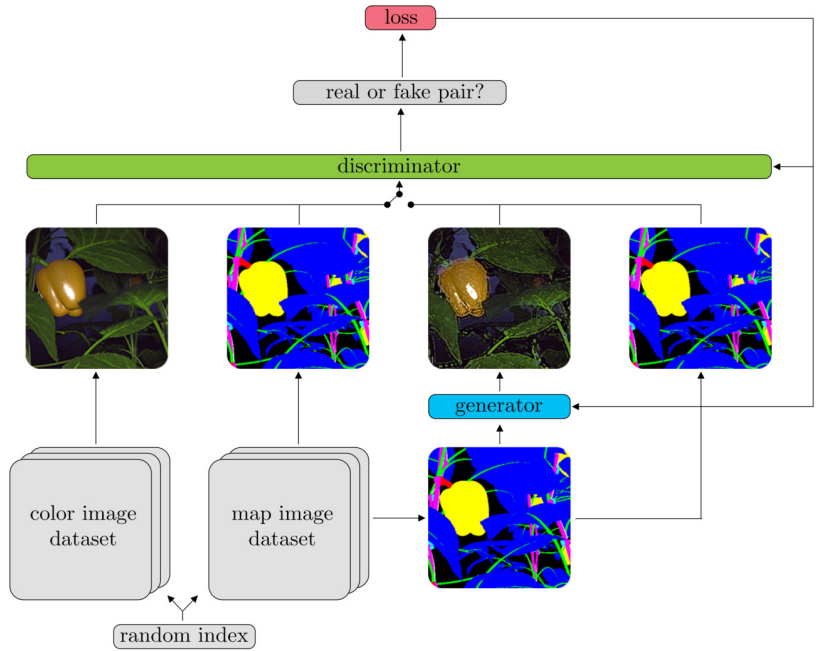

Figure 2: Learning schematic of a conditional generative adversarial network. In each learning step, a discriminator receives either a real pair of corresponding images in different domains (e.g. colored 3D render and a map) or a fake pair of which one domain image was translated by the generator (e.g. colored 3D render) from an image in the other domain (e.g. map). The discriminator determines if this image pair is real or fake. Through a loss function feedback to both the discriminator and the generator is given to optimise their tasks. In this example, the generator learns to translate render-like color images from class maps.

To achieve that, a cycle consistency loss was added to further regularise the learning [29]. Given a sample $x \in X$ and $y \in Y$, a loss was added to the optimisation such that $F(G(x)) \approx x$ and $G(F(y)) \approx y$. Hence, the learning was therefore constrained by the intuition that if an input image is translated from one domain to the other and then back again, an image should result similar to the original input. This similarity is captured by the cycle consistency loss, which forces the generators $G$ and $F$ to achieve unpaired geometrically consistent image-to-image translation from one domain to the other and vice versa.

In Figure 3 a schematic is shown of this learning process. Note that this method was pursued for this paper, because it allows to create a large dataset of images in the empirical domain. Furthermore, the key utility lies in the image pair P, in which the ground truth class mapping from the synthetic images could also be used for the translated synthetic images to the empirical domain. Moreover, the method does not require any annotations of empirical images.

The paper is structured in 2 parts, each with their corresponding materials, methods, discussion and conclusion sections. Part I describes and evaluates the image-to-image translation from the synthetic rendered domain to the empirical photographic domain. In Part II, the effect on segmentation learning was investigated using the translated images from Part I. The paper ends with a general discussion and conclusion.

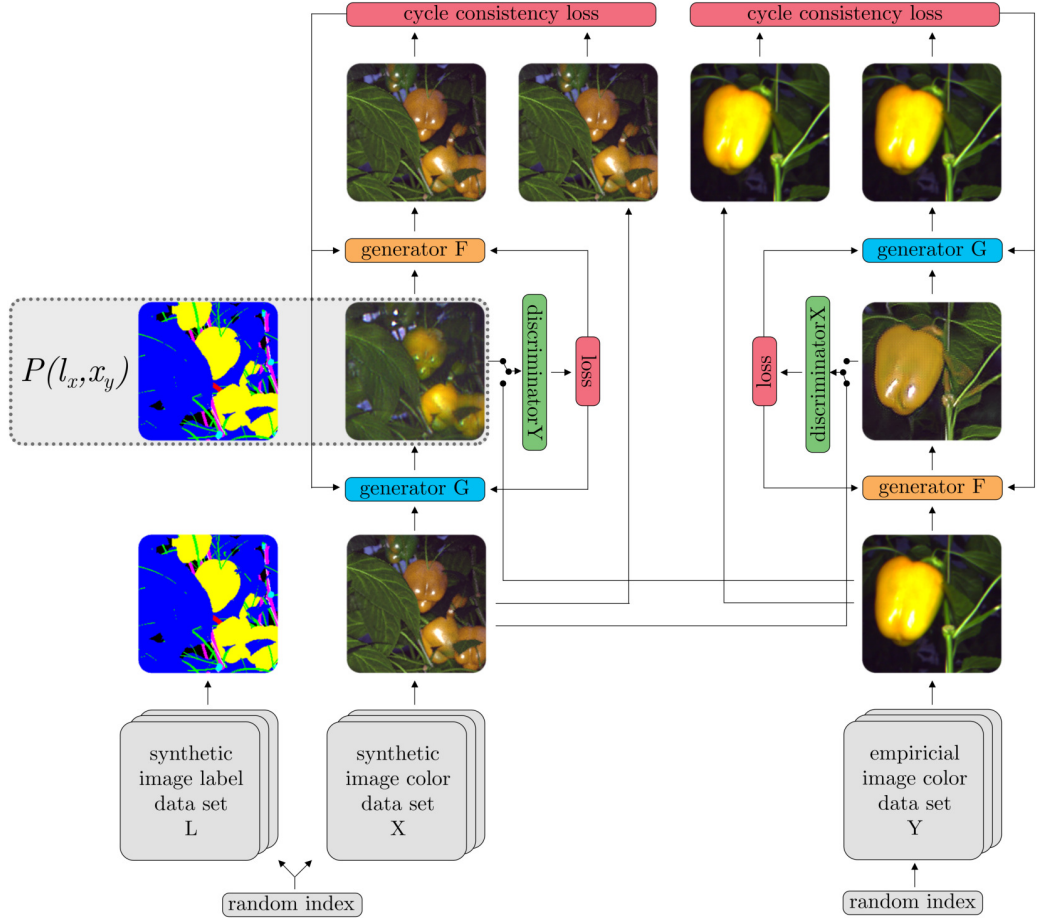

Figure 3: Learning schematic of a cycle generative adversarial network. In each learning step, generator G receives an image from domain X and generator F receives an image from domain Y. Each generator is trained to transform the input image to the other domain. A discriminator Y and discriminator X for each corresponding domain is trained to distinguish between generated and original domain images. From those first set of generated images, the opposing generator then synthesizes the second set of images back to its original domain (which ideally should result in the original domain image). A cycle consistency loss is then calculated by comparing the second set of images with the initial input image. The loss of both discriminators and cycle consistency is fed back to both generators for learning. In this example, each generator learns to synthesise an image to the opposing domain, whilst remaining geometrically consistent. This example was pursued in this paper to obtain image pair P, consisting of the label $l_x$ that corresponds to $x_y$; the translated image from domain X to Y.

## 2. Part I: Image-to-image translation

In this first part of the paper we describe and evaluate the unpaired image-to-image translation on agricultural images from the synthetic to the empirical domain and vice versa. The main objective was to obtain pairs of images P consisting of an image translated from the synthetic to the empirical domain, and a corresponding ground truth map (see Figure 3).

*2.1. Materials*

*2.1.1. Image dataset*

The unpaired image dataset [3] of *Capsicum annuum* (sweet- or bell pepper) was used that consists of 50 empirical images of a crop in a commercial high-tech greenhouse and 10,500 corresponding synthetic images, modelled to approximate the empirical set visually. In both sets, 8 classes were annotated on a per-pixel level, either manually for the empirical dataset or computed automatically for the synthetic dataset. In Figure 4 examples of images in the dataset are shown. The dataset was publicly released at:

`http://dx.doi.org/10.4121/uuid:884958f5-b868-46e1-b3d8-a0b5d91b02c0`

Both synthetic and empirical images were first cropped to 424x424 pixels to exclude the robot end-effector's suction cup in the image, because initial image-to-image translation experiments showed the cup was replicated undesirably in other parts of the image. This was in line with previous findings from the same authors where color and texture translation often succeeded, but domains with a large geometric variation were translated with less success [29]. Secondly, the image was resampled bilinearly to a resolution of 1000x1000 pixels as additional experiments during Part II showed upscaling improved the learning.

From the *Capsicum annuum* dataset, the synthetic images 1-1000 were used for training the translations and the remainder for testing. For the empirical images, 50 annotated images of the *Capsicum annuum* dataset were used for testing, whereas for training 175 non-annotated images were used that were not part of the released dataset, but were collected during the same data acquisition experiment.

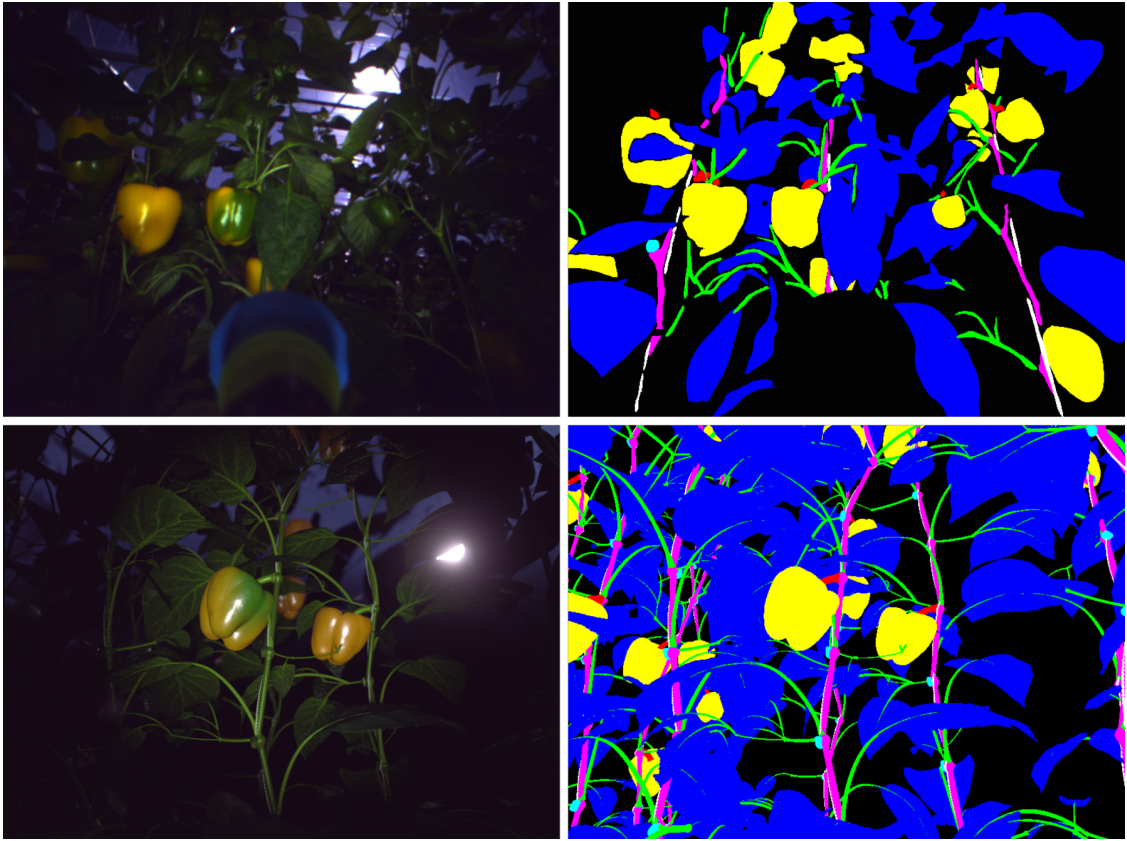

Figure 4: Uncropped examples of empirical (top row) and synthetic (bottom row) color images (left column) and their corresponding ground truth labels (right column). Part class labels: ● background, ● leafs, ● peppers, ● peduncles, ● stems, ● shoots and leaf stems, ○ wires and ● cuts where pepper where harvested.

*2.1.2. Software*

The Berkeley AI Research (BAIR) laboratory implementation of unpaired image-to-image translation using cycle-consistent adversarial networks was used [29].

*2.1.3. Hardware*

Experiments were run on a NVIDIA DevBox system with 4 TITAN X Maxwell 12GB GPUs, Intel Core i7-5930K and 128GB DDR4 RAM running Ubuntu 14.04.

*2.2. Methods*

The adversarial learning scheme in Figure 3 was applied with synthetic images as domain X and empirical images as domain Y. The hyper-parameters of the Cycle-GAN were manually optimised by visually evaluating the resulting images with their target domain. The number of generative and discriminative filters were set to 50 and the learning rate was set to 0.0002 with an ADAM [23] momentum term of 0.5. The basic discriminator model was used, whereas for the generator the RESNET 6 blocks model was used [19]. Weights for the cycle loss were set to 10 for each translation direction.

*2.2.1. Quantitative translation evaluation*

Although the success of the translation is already quantitatively captured by the adversarial loss, this measure is biased and mathematically obfuscated. By specifically looking at key image features like color, contrast, homogeneity, energy and entropy, it could be derived if the translated images improved on those features. This would provide evidence about the dissimilarity gap between the synthetic and empirical domains.

For this purpose, we first compared for each object part class the synthetic color distribution prior and post translation with those of the empirical distribution. The color spectrum of each class was obtained by first transforming the color images to HSI colorspace. The Hue channel in the transformed image represented for each pixel which color was present, regardless of illumination and saturation intensity. The histogram of this channel was then taken to count the relative color occurrence per class.

As we hypothesise that the color difference between the synthetic and the empirical domain images will be reduced after translation of the synthetic images, the correlations of the color distributions of each object part class were compared for i) the empirical images and the synthetic images, and ii) the empirical images and the translated images.

Second, to obtain additional image features, first an average gray level co-occurrence matrix (GLCM) [16]was calculated for each class for the first 10 images in the synthetic, synthetic translated to empirical and empirical sets. The GLCM summarises how often a pixel with a certain intensity value $i$ occurs in a specific spatial relationship to a pixel with the intensity value $j$. This relationship was set to address horizontally neighbouring pixels only. From the GLCM, the following features were derived:

**Contrast** $= \sum_{i,j} |i-j|^2 \, GLCM(i,j)$, measuring the overall difference in luminance between neighbouring pixels.

**Homogeneity** $= \sum_{i,j} \frac{GLCM(i,j)}{1+|i-j|}$, a value that measures the closeness of the distribution of elements in the GLCM to the GLCM diagonal, which implies that high values of homogeneity reflect the absence of changes in the image and indicates a locally homogenous distribution in image textures.

**Energy** $= \sum_{i,j} GLCM(i,j)^2$, a measure of texture crudeness or disorder.

**Entropy** $= \sum_{i,j} -ln(GLCM(i,j)) \cdot GLCM(i,j)$, measuring the amount of information or complexity in the image.

*2.3. Results*

In Figure 5 the results of the image-to-image translations are shown. The second column is of most interest to our research, as it shows the set $X_y$ of synthetic images which were translated to the empirical domain. However, as a reference also the translation from empirical to the synthetic domain is shown in the third column.

The color distributions for each object part class for the synthetic, empirical and translated synthetic images are shown in Figure 6. The corresponding correlations between the empirical images and the synthetic as well as translated synthetic images are shown in Table 1.

For the image features contrast, homogeneity, energy and entropy, the results per class for the synthetic, empirical and synthetic translated to empirical images are shown in Figure 7. The difference of 0.100 was found in contrast averaged over all classes between the synthetic and empirical set, whereas this difference was reduced to 0.015 for the translated and the empirical set. Similarly, for homogeneity this was reduced from 0.028 to 0.015. For the energy feature, this was reduced from 0.126 to 0.026. Regarding entropy, the average difference was reduced from 0.364 to 0.003.

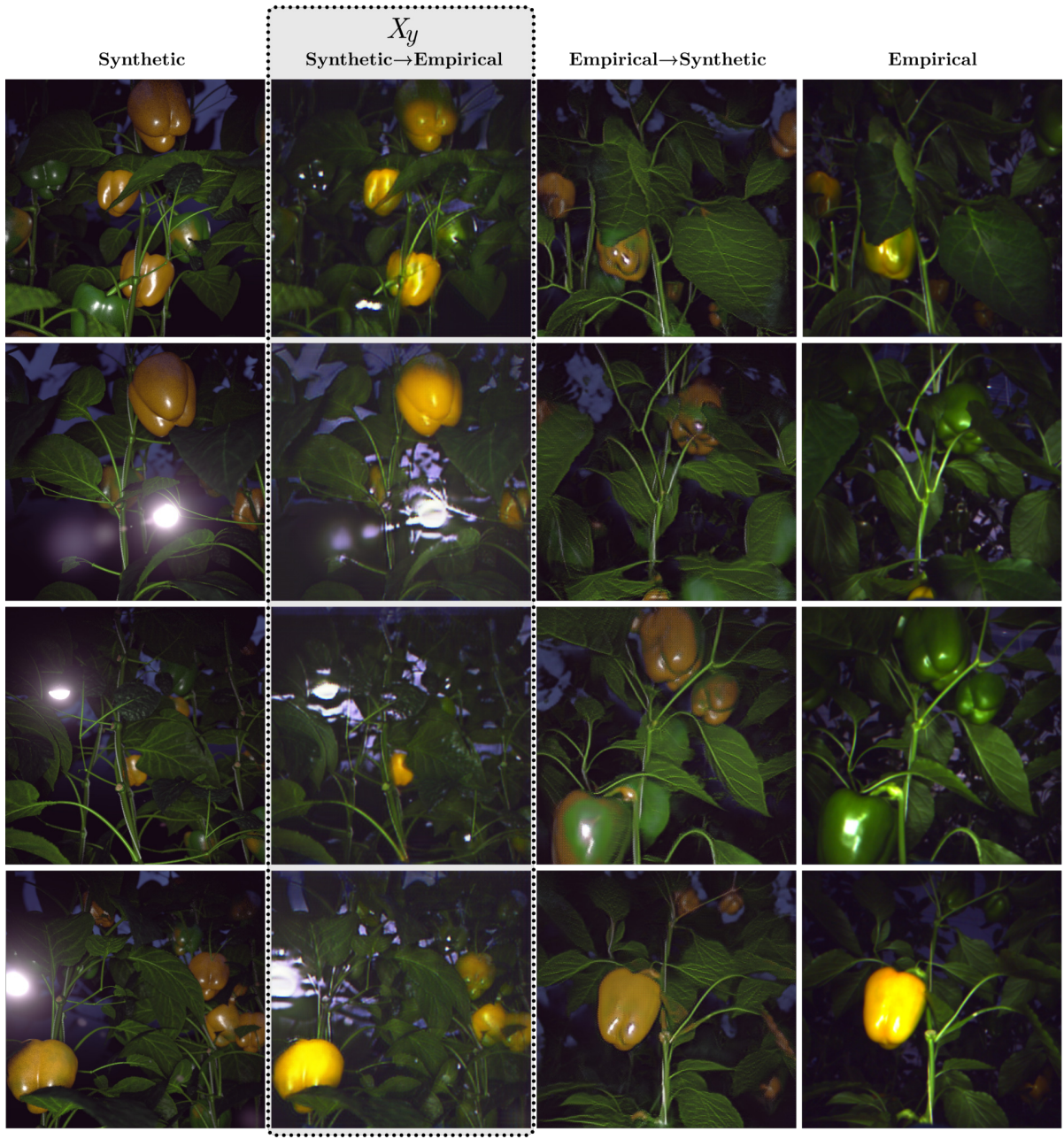

Figure 5: Image-to-image translation examples using Cycle-GAN. Source domain images prior translation are shown in the outer columns; synthetic images (left) and empirical images (right). The second column shows the set of interest $X_y$; the translated synthetic images to the empirical domain. The third column shows empirical images translated to synthetic domain.

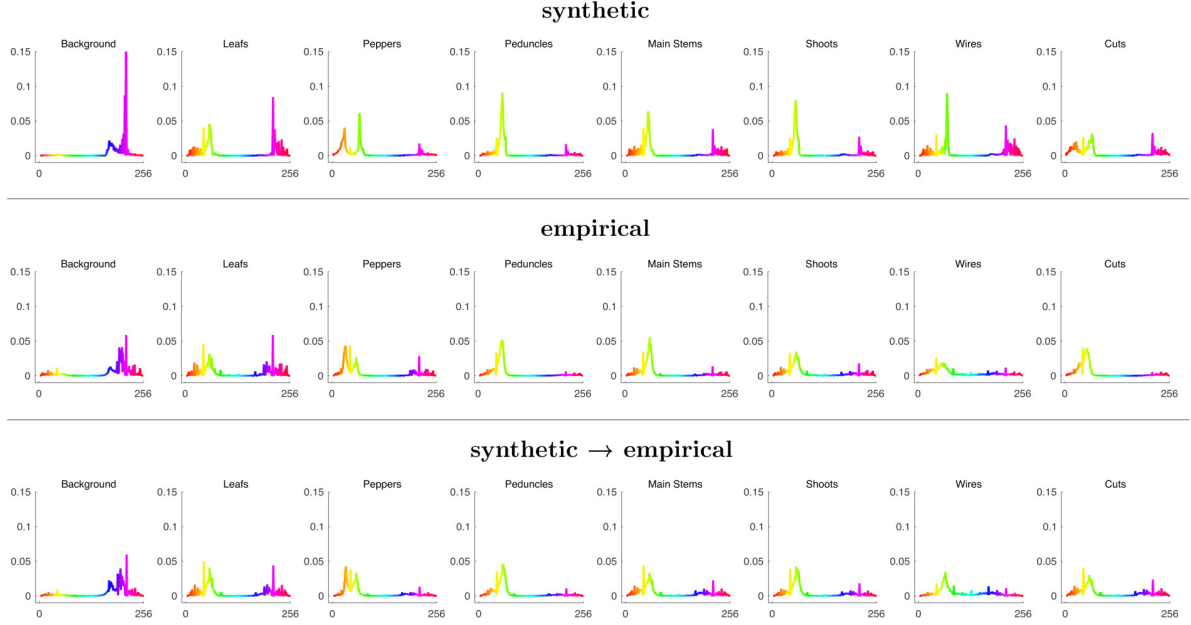

Figure 6: Color distributions discretized to 256 values in the hue channel (x-axis) per class of the synthetic, empirical and synthetic translated to empirical images. Integral per distribution amounts to 1 (y-axis).

|  | backgr. | leafs | peppers | peduncles | stems | shoots | wires | cuts | mean |
|---|---|---|---|---|---|---|---|---|---|
| **correlation(synthetic, empirical)** | 0.25 | 0.78 | 0.42 | 0.93 | 0.76 | 0.83 | 0.45 | 0.48 | **0.62** |
| **correlation(synthetic→empirical, empirical)** | 0.86 | 0.94 | 0.93 | 0.93 | 0.92 | 0.98 | 0.81 | 0.79 | **0.90** |

Table 1: Average color distribution correlations per object part class between i) the empirical images and synthetic images, and ii) the empirical images and the translated synthetic images.

*2.4. Discussion and conclusion*

Qualitative visual evaluation of the results in Figure 5 showed a remarkable translation of synthetic images to empirical looking images and vice versa. Most notably the scattering of illumination and color of each plant part were converted realistically. It also appeared that the model learns to distinguish plant parts without any supervised information, as the (partially) ripe and unripe fruit were often translated to the other domain with altered maturity levels. A difference in camera focus seemed translated properly, indicating that local features (e.g. edge blur and texture) can be mapped accurately.

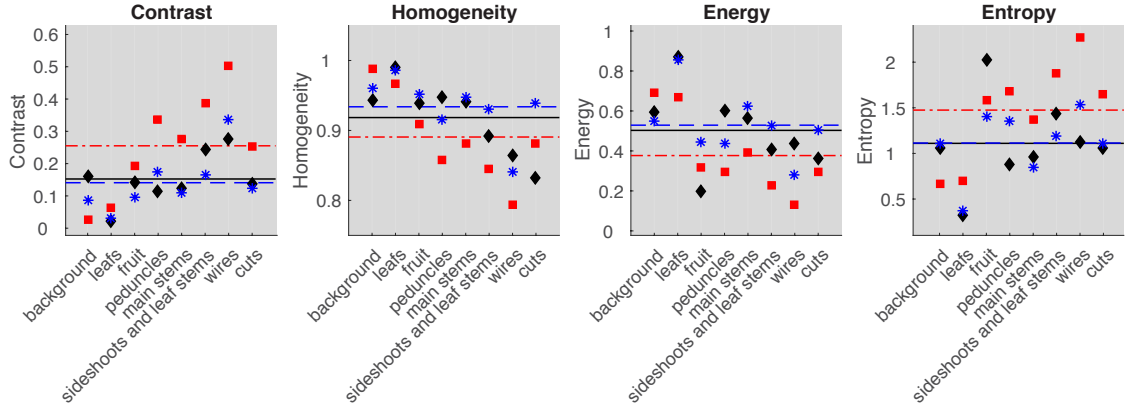

Figure 7: Image features values for contrast, homogeneity, erergy and entropy per class for the empirical ◆, synthetic ■ and synthetic translated to empirical ∗ images. Average over all classes is represented by a solid line for the empirical set, a dashed-dotted line for the synthetic set and a dashed line for the synthetic translated to empirical set.

Some image artifacts did arise however, especially the translation of overexposed areas like sunshine or fruit reflections. The explanation might be that the model cannot generate this information correctly because any information beyond overexposure prior translation was already collapsed to a single maximum value (e.g. 255). Furthermore, an overlay of a checkerboard-like texture seems to have been added to the translated local textures. The image-to-image translation method appeared not to be suited when one image set contained additional objects or parts that were absent in the other set, such as the presence of a suction cup in our earlier experiments. We noticed in previous experiments that this part was undesirably replicated in other areas of the image.

In Figure 5 we can also see that large morphological features (e.g. large plant part shape and geometry relatively to other plant parts) were not translated, indicating a limitation of the Cycle-GAN approach. However, since geometry was not translated, this did allow for using the underlying synthetic ground truth labels to be used with the translated images for Part II. If also the geometry would have been translated, then the ground truth labels would not have been translated accordingly.

In Figure 6, the translation effect on color distribution can be seen for each plant part and background. Quantitatively, the mean color correlation of 0.62 between the synthetic and empirical images increased post translation to 0.90 (See Table 1 for correlations per plant part and the mean over all plant parts). Indeed this is also what we observe in Figure 5, where for example the color of the fruit in the translated synthetic images matches the empirical images more than those of the synthetic images.

When we look at the averages of the image texture features contrast, homogeneity, energy and entropy, they were closer together when comparing the empirical images and synthetic translated images than when comparing the empirical images and the synthetic images. For some individual classes this did not hold however, e.g. the homogeneity of the cuts was erroneously doubled instead. In Figure 5 it can indeed be observed that local level textures of the translated synthetic images have become more similar to those of the empirical images. For example, the smoothness of the fruit in the translated synthetic images is improved towards the empirical images, as compared to the the more coarse and grainy surface texture of the fruit in the synthetic images.

Regarding our first hypothesis, we therefore confirm that image feature differences with the empirical set were reduced after translation of the synthetic images, using a cycle-GAN.

This part of the work contributed to the field of computer vision (e.g. for agricultural robotics) by providing a method for optimising realism in synthetic training data to potentially improve state-of-the-art machine learning methods that semantically segment plant parts, as evaluated in Part II of this paper.

## 3. Part II: Improving semantic segmentation

In Part II, the effect on using translated images on object part segmentation learning was investigated by using the translated images from Part I instead of synthetic images. Our second hypothesis states that by bootstrapping with translated images and empirical fine-tuning, the highest empirical performance can be achieved over methods that bootstrap with limited dataset size of (30) empirical images or a large set (8750) of synthetic images. With our third hypothesis in this paper, we reckon that without any empirical fine-tuning, learning can be improved with translated images as compared to using only synthetic bootstrapping.

*3.1. Materials*

The synthetic and empirical datasets as described in Part I (see Section 2.1.1) were used as well as the obtained image pairs $P(l_x, x_y)$ (see Figure 2).

*3.1.1. Software*

The publicly available semantic segmentation framework DeepLab V2 was used, which implemented convolutional neural network (CNN) models [26, 7] on top of Caffe [22]. Specifically, the VGG-16 network was used with a modification to include *à trous* spatial pyramid pooling for image context at multiple scales by convolutional feature layers with different fields-of-view [8, 18].

*3.1.2. Hardware*

Experiments were run on the same hardware as used in Part I. As a dependency for the DeepLab V2 Caffe version, the archived version of CUDA 7.5 was installed.

*3.2. Methods*

To compare performance differences, 7 experiments were performed using different combinations of train, fine-tune and test sets. The motivation for each experiment is given below and the used sets and image ranges are shown between brackets.

**A** *Train: empirical (1-30). Test: empirical (41-50).*

An experiment to see if the model can learn using only a small empirical dataset. This provides a reference for comparison of performance with other experiments that bootstrap with synthetic or translated synthetic images and/or fine-tune with empirical images. Given the small training size of the dataset in this experiment, the performance was expected to be low, compared to all other experiments that tested on empirical images.

**B** *Train: synthetic (1-8750). Test: synthetic (8851-8900).*

This experiment was run to obtain baseline performance of the model when having access to a large and detailed annotated synthetic dataset. Performance is expected to be highest of all experiments because of the perfect labels, largest dataset size and relatively low image feature variance compared to empirical or synthetic translated images.

**C** *Train: synthetic (1-8750). Test: empirical (41-50).*

A reference experiment to see to what extent a network trained on synthetic images can generalise to the empirical domain, without fine-tuning with empirical images. Given the similarity gap between synthetic and empirical data, the performance should be relatively low compared to that of Experiment A or when compared to experiments that trained on a more realistic dataset, e.g translated synthetic as in Experiment F.

**D** *Train: synthetic (1-8750). Fine-tune: empirical (1-30). Test: empirical (41-50).*

Similar to Experiment C, but with an extra fine-tuning step using empirical images. Performance is expected to be higher than C, because the network also optimises for the empirical image feature distribution. The performance of this experiment is expected to be lower than that of Experiment G, where the synthetic images were replaced by translated synthetic images, because the synthetic image feature distribution is more dissimilar with the empirical distribution than the translated synthetic distribution with the empirical distribution is.

**E** *Train: synthetic translated to empirical (1-8750). Test: synthetic translated to empirical (8851-8900).*

This experiment was run to obtain baseline performance of the model when having access to a large and detailed annotated translated synthetic dataset. The performance should be similar of that of Experiment B, though is expected to a bit lower due to the extra variance that the empirical feature distribution might have introduced when synthetic images were translated to the empirical domain.

**F** *Train: synthetic translated to empirical (1-8750). Test: empirical (41-50).*

With this experiment, we could check to what extent a synthetic trained network with improved realism can generalise to the empirical domain, without fine-tuning with empirical images. This experiment should provide the main result for our third hypothesis that states that without any fine-tuning with empirical images, improved learning for empirical images can be achieved using only translated images as opposed to using only synthetic images, as evaluated in Experiment C.

**G** *Train: synthetic translated to empirical (1-8750). Fine-tune: empirical (1-30). Test: empirical (41-50).*

This experiment should provide the main result for our second hypothesis, that states the synthetic images translated to the empirical domain can be used for improved learning of empirical images, as compared to using only synthetic images for bootstrapping (Experiment D). Performance of this experiment was expected to be the highest amongst all our experiments that tested on empirical data, because a large dataset with high similarity with the empirical images was used in combination with fine-tuning on empirical images.

*3.2.1. CNN Training*

For each experiment, a convolutional neural network was trained and/or fine-tuned and tested according to the dataset scheme as described in Section 3.2.

The hyperparameters of the network were manually optimised using separate validation datasets for combination of models and data set configurations as suggested by [14, 5]. This resulted in using Adaptive Moment Estimation (ADAM) [23] with $\beta_1 = 0.9$, $\beta_2 = 0.999$, $\varepsilon = 10^{-8}$ and a base learning rate of 0.001 for 30,000 iterations with a batch size of 4. These chosen hyper-parameters were found to be consistently optimal previously [4] and therefore we fixed them across conditions. An adjustment was made in the layer weight initialisation procedure, by updating the model to using MSRA weight fillers [20, 25]. Furthermore, the dropout rate [28] was adjusted to 0.50 to circumvent early overfitting and facilitate generalisation. The size of the input layer was cropped to 929x929 pixels, which was the maximum that our GPU memory could handle.

*3.2.2. Performance Evaluation*

To calculate the performance of the segmentation, we used the Jaccard Index similarity coefficient as an evaluation procedure, also known as the intersection-over-union (IOU) [17] which is widely used for semantic segmentation evaluation [12, 11]. This measure is defined in Equation 2, where the mean IOU over all object part classes equals the intersection of the segmentation and the ground truth divided by their union. A higher IOU implies more overlap, hence better performance. To derive the measure, a pixel-level confusion matrix C was calculated first for each image $I$ in data set $D$ :

$$C_{ij} = \sum_{I \in D} \left| \{ p \in I \mid S_{gt}^I(p) = i \wedge S_{ps}^I(p) = j \} \right|, \qquad (1)$$

where $S^I_{gt}(p)$ is the ground truth label of pixel $p$ in image $I$ and $S^I_{ps}(p)$ is the predicted segmentation label. This implies that $C_{ij}$ equals the number of predicted pixels $i$ with label $j$. The average IOU over all classes $L$ is given by:

$$IOU = \frac{1}{L} \sum_{i=1}^{L} \frac{C_{ii}}{G_i + P_i - C_{ii}} \; , \; \text{where} \quad (2)$$

$$G_i = \sum_{j=1}^{L} C_{ij} \; \text{ and } \; P_j = \sum_{i} C_{ij} \quad (3)$$

Hence $G_i$ denotes the total number of pixels labeled with class $i$ in the ground truth and $P_j$ the total number of pixels with prediction $j$ in the image.

*3.3. Results*

In Figure 8 the average IOU over all classes for Experiments A through G is shown, as well as previous results of similar experiments A-D [4]. In Figure 9 the performances were split over the object part classes. Qualitative results are presented in Figure 10.

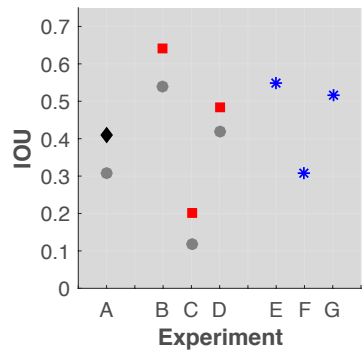

Figure 8: Average IOU over all object part classes for Experiment A with empirical training (◆), Experiments B,C and D with synthetic image bootstrapping (■) and Experiments E, F and G with synthetic translated to empirical image bootstrapping (✻). Previous perfomance [4] for similar experiments A-D are shown for reference (●).

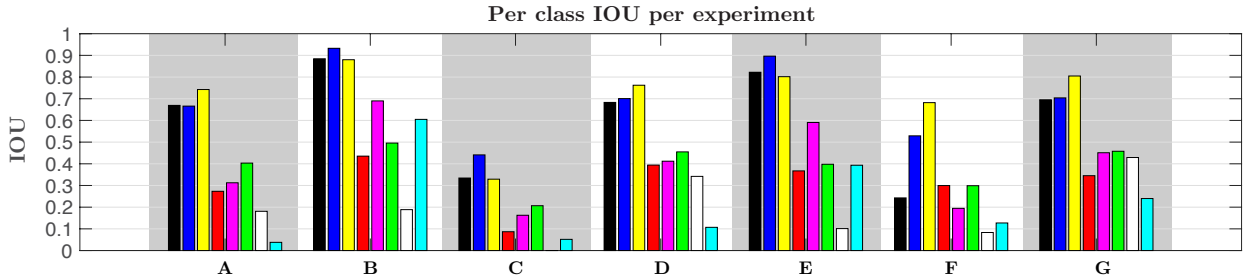

Figure 9: For Experiments A through G, the IOU per class is displayed, ordered as: ● background, ● leafs, ○ peppers, ● peduncles, ● stems, ● shoots and leaf stems, ○ wires and ○ cuts where pepper where harvested.

*3.4. Discussion and conclusion*

Compared to the former attempts using the same dataset [4], current results showed an overall improved performance of 0.08 IOU on average for Experiments A through D, as can be seen in Figure 8. The two differences implemented in our current attempt were first the cropping to 424x424 pixels to exclude the suction cup and then the upscaling to 1000x1000 pixels. The same CNN configuration was used. Additional experiments showed that the upscaling was the main cause of the performance increase. This might be explained by the CNN's larger field of view, allowing for detail only to dissolve by convolutions and pooling in deeper layers of the network.

In Experiment A, the aim was to see if the model can learn to segment empirical images using only a small empirical training dataset. The CNN reached a performance of 0.41 average IOU (see Figure 8). Relative to the other experiments testing on empirical images, the performance was expected to be low due to the small training dataset size. Indeed compared to Experiment D and G the performance of Experiment A was lower, which might be caused by the D and G models being bootstrapped on synthetic or translated synthetic images. However, the performance of Experiment A was higher compared to Experiments C and F, which can be explained by those experiments not using any empirical data during training.

Looking at the per class performance distribution of Experiment A in Figure

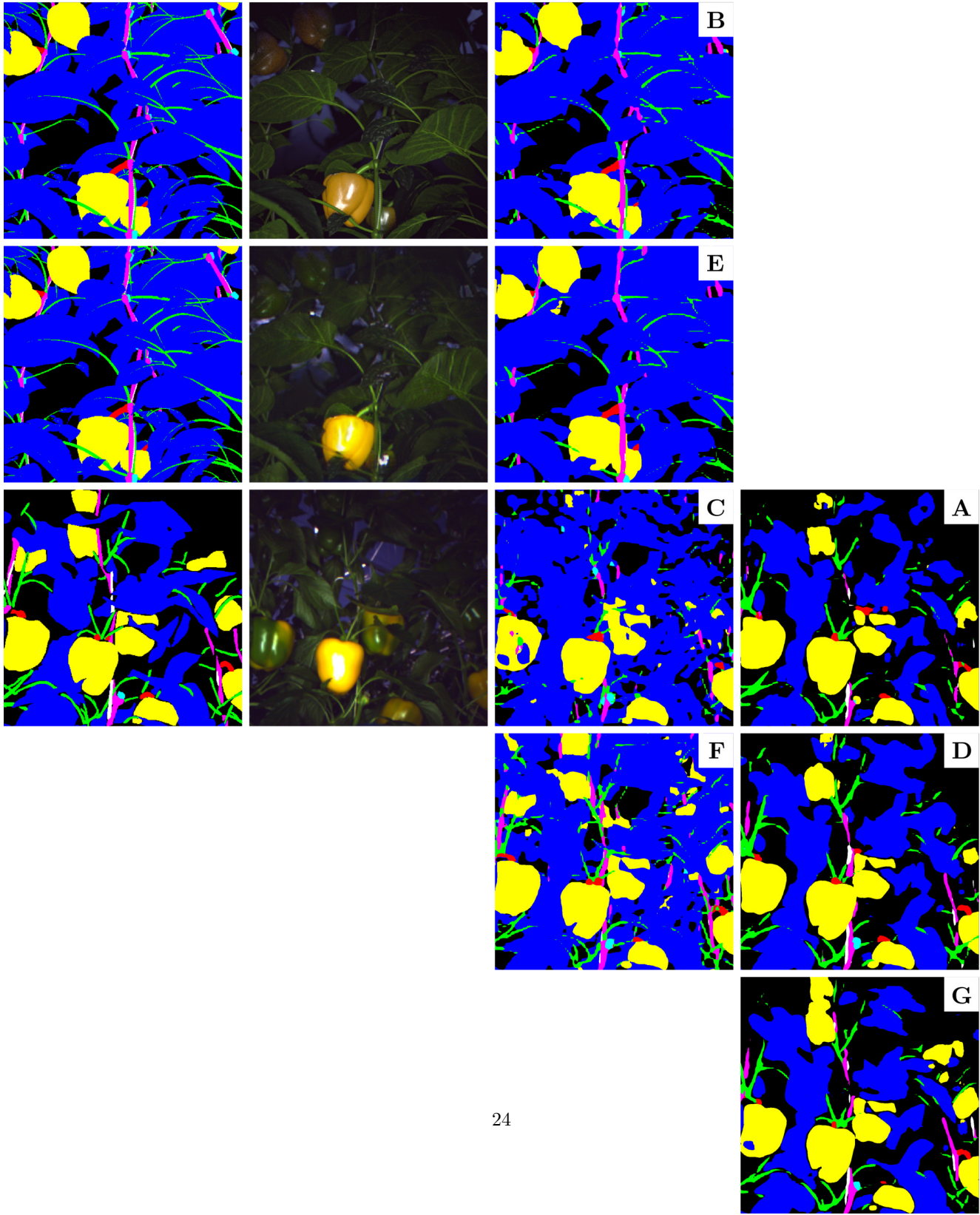

Figure 10: Qualitative results from Experiments A through G. In the first column, the ground truths for synthetic (top), synthetic translated to empirical (middle) and empirical (bottom) are shown with labels: ● background, ● leafs, ● peppers, ● peduncles, ● stems, ● shoots and leaf stems, ○ wires and ● cuts. In the second column color images are displayed. Experimental results are grouped in the third column (trained without empirical data) and fourth column (fine-tuned with empirical data).

9, we note that the *cut* class was barely recognised having an IOU of 0.04. Recognising all classes was previously considered as a requirement [4]. Therefore we concluded that training with empirical images alone did not suffice, although qualitative results (see Figure 10) looked promising and useful for some tasks like fruit detection.

Experiment B was run to obtain baseline performance of the model when having access to a large and detailed annotated synthetic dataset. Performance was expected to be highest of all experiments because of the perfect labels, largest dataset size and relatively low image feature variance compared to empirical or synthetic translated images. Indeed B achieved the best performance with an average IOU of 0.64. This performance could be used as a baseline to indicate the maximum obtainable IOU for this domain and currently used CNN architecture. The performance of the other experiments should be put into perspective of this IOU. Qualitative results still showed some gaps in thin and elongated classes like leaf stems and shoots, although results were much improved over previous segmentations where such gaps were larger [4].

Experiment C was a reference experiment to see to what extent a network trained on synthetic images can generalise to the empirical domain, without fine-tuning with empirical images. With an average IOU of 0.20, the performance approximately doubled over previous 424x424 results [4]. Furthermore, the performance was lower than that of Experiment A, likely because the used synthetic training data was too dissimilar with the empirical images.

Looking at the per class distribution, classes like *peduncle* and *cut* were barely recognised (IOU<0.1) and the *wire* class was omitted all together. Qualitatively, results looked far from similar to the ground truth. Looking at the qualitative results, we can conclude that training only with synthetic data would not be sufficient for many tasks, given the current learning architecture.

In Experiment D, a similar training scheme of Experiment C was used, but with an extra fine-tuning step using empirical images. The IOU performance on empirical images was increased to 0.48. Hence bootstrapping with synthetic images increased performance by 17% over no bootstrapping in Experiment A.

We conclude that bootstrapping with synthetic images and fine-tuning with empirical images can be used to close the gap towards the optimal estimated possible performance of Experiment B. Furthermore, we note that all classes were included, although the *cut* class was again barely recognised (IOU=0.11). Qualitatively, results looked close to the ground truth.

Experiment E trained and tested on a large dataset, similar to Experiment B, but instead of synthetic images the translated synthetic images were used. The performance of 0.56 IOU was lower than of the IOU=0.64 from Experiment B, as expected and probably due to the extra variance that the empirical feature distribution might introduce when synthetic images were translated to the empirical domain. Qualitatively results looked comparable.

Experiment F evaluated on empirical images when trained on synthetic translated to empirical images, without fine-tuning with empirical images. With this experiment, we could check to what extent a synthetic trained network with improved realism can generalise to the empirical domain, without yet fine-tuning with empirical images. Compared to Experiment C (using synthetic images instead of translated ones), the performance increased with 55% to an average IOU of 0.31. This experiment confirms our third hypothesis that without any fine-tuning with empirical images, improved learning for empirical images can be achieved using only translated images as opposed to using only synthetic images. Although qualitatively, also improvements could be observed over Experiment C, we noted from the class performance distribution there existed still a relative poor performance on the classes *wires* and *cuts*.

In Experiment G, the model from Experiment F was fine-tuned with empirical images. This experiment should provide the main result for our second hypothesis, that states that synthetic images translated to the empirical domain can be used for improved learning of empirical images, as compared to using only synthetic images for bootstrapping, as evaluated in Experiment D. Our hypothesis is confirmed by achieving the best performance on empirical data of an IOU=0.52. This was an increase of 27% over Experiment A (only training on empirical images) and 8% over Experiment D.

Qualitatively, results in Experiment G looked close to the ground truth and comparable to results of Experiments A and D. Looking at the class distribution, all classes were included. Most notably the *cut* class performance increased with 118% over Experiment D and with 600% over Experiment A to an IOU of 0.24.

To summarise, we have seen that without using any annotated empirical training images, an improved performance can be achieved by bootstrapping with translated synthetic images instead of synthetic images. Furthermore, we have shown that by also fine-tuning with a small empirical dataset, the highest performance on empirical images can be achieved.

## 4. General discussion and conclusion

In Part I, a cycle consistent generative adversarial network was applied to synthetic and empirical images with the objective to generate more realistic synthetic images by translating them to the empirical domain. Our analysis showed that the image feature distributions of these translated images, both in color and texture, were improved towards the empirical images. Regarding our first hypothesis, it was confirmed that the image feature difference with the empirical set was reduced after translation of the synthetic images. Qualitatively, the translated synthetic images looked highly similar to the real world images. However, some translation artifacts appeared. Furthermore the Cycle-GAN method could not improve upon geometric dissimilarities between the synthetic and the empirical domain. The latter proved an advantage however, as the synthetic ground truth also corresponded to the translated color images, allowing for the experiments on improved learning in the second part of our work.

In Part II, it was evaluated to what extent translated synthetic images to the empirical domain could improve on CNN learning with empirical images over other learning strategies. We confirmed our second hypotheses that by using translated images and fine-tuning with empirical images, the highest performance for empirical images can be achieved (IOU=0.52) compared to training with only empirical (IOU=0.41) or synthetic data (IOU=0.48)

Besides improving segmentation performance on empirical images using translated synthetic images instead of only empirical or synthetic images during training, another key contribution of our work is the further minimisation of the CNN's dependency on annotated empirical data. We confirmed our third hypothesis that without any empirical image fine-tuning, learning can be improved with translated images (IOU=0.31), a 55% increase over just using synthetic images (IOU=0.20).

The work presented in this paper can be seen as an important step towards improved sensing for applied computer vision domains such as in agricultural robotics, medical support systems or autonomous navigation. It facilitates CNN semantic object part segmentation learning without or minimal requirement of annotated images.

**Acknowledgement**

This research was partially funded by the European Commission in the Horizon2020 Programme (SWEEPER GA no. 644313) and the Dutch Ministry of Economic Affairs (EU140935).